\documentclass{ecai}
\usepackage{times}
\usepackage{graphicx}
\usepackage{latexsym}

\usepackage{balance}

\newcommand{\tax}{\mathit{ax}}

\begin{document}

\title{
%
	On Expert Behaviors and Question Types for Efficient Query-Based Ontology Fault Localization\thanks{Extended versions of this paper were published in the Proceedings of the \emph{International Conference on Industrial, Engineering and Other Applications of Applied Intelligent Systems (IEA/AIE-2019)} \protect\cite{rodler2019usefulness} and in the \emph{Knowledge-Based Systems (KBS)} journal \protect\cite{rodler2022kbs}. Please find (self-explaining) slides of this work at http://tiny.cc/b00igz.}
%
}

\author{Patrick Rodler\institute{University of Klagenfurt, Austria, email: patrick.rodler@aau.at} } 

\maketitle
\bibliographystyle{ecai}

\begin{abstract}
We challenge existing query-based ontology fault localization methods wrt.\ assumptions they make, criteria they optimize, and interaction means they use. We find that their efficiency depends largely on the behavior of the interacting expert, that performed calculations can be inefficient or imprecise, and that used optimization criteria are often not fully realistic. As a remedy, we suggest a novel (and simpler) interaction approach which overcomes all identified problems and, in comprehensive experiments on faulty real-world ontologies, enables a successful fault localization while requiring fewer expert interactions in 66\,\% of the cases, and always at least 80\,\% less expert waiting time, compared to existing methods.
\end{abstract}

\section{Introduction, Discussion and Approach}

\noindent\textbf{Motivation.} As Semantic Web technologies have 
become widely adopted in, e.g., 
security
and health applications, the quality assurance of the 
knowledge 
underpinning
these applications is a critical requirement. At the core of 
semantic technologies, ontologies are a means to represent knowledge in a formal, structured and human-readable way, with a 
well-defined semantics. 
Due to high ontology complexity,
expressive logics used, or distributed, collaborative and tool-supported 
development processes pursued, faults in ontologies are frequent \cite{copeland2013finding,meilicke2007repairing}. 
Among those, faults that affect the ontology's semantics (e.g., 
wrong entailments) are of particular concern, 
e.g., a medical system could suggest a wrong therapy for a patient. Since manual quality assurance is virtually infeasible for present-day ontologies, a range of tools have been proposed, aiming i.a.\ at fault prevention \cite{rector2004owl,RousseyCB09}, detection \cite{copeland2013finding}, localization \cite{meilicke2007repairing,schlobach2007debugging} and repair \cite{horridge2009explaining,Kalyanpur2006a,troquard2018repairing}.

\noindent\textbf{(Query-based) Ontology Fault Localization.}
This work is devoted to the problem of fault localization, based on ideas from the field of 
\emph{model-based diagnosis} \cite{dekleer1987,Reiter87}: 
Given an ontology that violates predefined \emph{requirements} (e.g., consistency, coherency, no unwanted entailments),
find a \emph{diagnosis}, i.e., an irreducible set of axioms whose faultiness explains all requirements violations. A deletion or adequate modification of the axioms of a diagnosis leads to a repaired ontology compliant with the given requirements. 
However, a sub-problem inhering fault localization is the fact that there is often a substantial number of competing diagnoses, where all of them lead to repaired ontologies with different semantics \cite{Rodler2015phd}. 
Hence, identifying the \emph{actual diagnosis} (the one diagnosis pinpointing the \emph{actually} faulty axioms, which leads to a repaired ontology with the intended semantics)
is a pivotal step towards meaningful and sustainable ontology repair. 
One particularly appealing \cite{rodler2019KBS_userstudy} approach to this end is \emph{interactive query-based ontology debugging} \cite{Rodler2015phd,Shchekotykhin2012}, where additional information to isolate the actual diagnosis is gathered by an interactive system in terms of a query-answer dialog with a \emph{(domain) expert}. Each \emph{query} is a true/false-question about (non-)entailments of the intended ontology and has the property to invalidate at least one diagnosis regardless of the answer. An example of a query 
from a medical domain would be $Q=\{\mathsf{Tumor} \sqsubseteq \exists \mathsf{causes}.\mathsf{Pain}\}$,
i.e., 
``Does every tumor cause pain?'' Queries and their associated answers are
used by the system as \emph{test cases} to successively rule out diagnoses, until ultimately a single (highly probable) one remains. This interactive technique is especially attractive as it lets the user focus on the intended ontology rather than on (the analysis of) the faulty one. More specifically, it relieves the expert from the need to analyze which or why entailments do (not) hold \emph{in the faulty ontology} or why 
\emph{the faulty ontology} does not meet the given requirements. Instead, the task of the expert reduces to the mere classification of certain axioms as either entailments
or non-entailments 
\emph{of the intended ontology}.
 
\noindent\textbf{Challenges and Goals.}
%
%
Since expert consultations are costly,
query-based debuggers pursue the following \emph{goals}:
\emph{(G1)}~Find the actual diagnosis \emph{(G2)}~with the least effort and \emph{(G3)}~with the least waiting time for the interacting user.
%
%
%
The following \emph{influencing factors} determine how well these goals can be approached: \emph{(F1)}~The way of interacting with the expert (\emph{how to define a query?}), \emph{(F2)}~the expert behavior (\emph{how will the expert answer queries?}), \emph{(F3)}~the criterion to be optimized (\emph{how to measure the expert's effort?}), and \emph{(F4)}~the used algorithm for query computation 
(\emph{how to compute the best query?}).

\noindent\textbf{Existing Approaches.} 
We now discuss how existing query-based methods address these questions: 
(F1)~A \emph{(normal) query} is \emph{a set of} axioms $Q=\{\tax_1,\dots,\tax_k\}$ (this definition is natural for algorithmic and computational reasons, cf.\ \cite{rodler2019usefulness} for details). 
Showing the expert $Q$ means asking them whether or not $\tax_1\land\dots\land \tax_k$ is entailed by the intended ontology. 
%
%
(F2)~The 
(\emph{query-based}) \emph{expert} is viewed as a function that maps \emph{queries} to either \emph{true} or \emph{false}, where \emph{true} (\emph{false}) means that each (some) query-axiom is (not) entailed by the intended ontology. Clearly, to answer positively, the expert must examine each query-axiom; 
as opposed to the negative case, where it suffices to know some non-entailed query-axiom.
%
%
So, whether (and how much) information beyond the mere answer \emph{false} (i.e., that some \emph{undefined} query-axiom is non-entailed) is obtained depends on the expert at hand. 
%
To study the impact of different answering behaviors on fault localization efficiency,
we complement the (existing) notion of the query-based expert with the one of an \emph{axiom-based expert}, i.e., a function which maps \emph{query-axioms} to either \emph{true} or \emph{false}. 
While query-based and axiom-based experts are equally-behaving in the affirmative case, we can conceive of various axiom-based sub-types in the negation case,
e.g., the \emph{minimalist} (classifies one query-axiom by false), the \emph{pragmatist} (classifies query-axioms one-by-one, until and including the first negative one that is encountered), and the \emph{maximalist} (classifies each query-axiom). Note that each (negative) axiom-based answer is strictly more informative than a query-based one, i.e., an axiom-based answering strategy 
means better diagnoses elimination and less cost.
(F3)~Most often, the \emph{number of queries (\#Q)} is used to quantify fault localization cost. Because the (global) minimization of \#Q
is NP-hard, 
query selection \emph{heuristics} \cite{rodler17dx_activelearning,rodler2018ruleML,Shchekotykhin2012}
are employed for choosing the best query in each iteration based on a (local) one-step-lookahead assessment (how favorable is the expected situation after a query is answered?). 
However, these heuristics do not take into account the \emph{number of axioms (\#Ax)} an expert has to classify, although the size of different queries in terms of the included axioms can vary considerably. Hence, we argue that \#Ax is a more realistic measure to evaluate the expert's effort. 
%
%
Moreover, a query defined as a set of axioms, see (F1), coupled with the fact that 
heuristics perform a binary
(\emph{true} vs.\ \emph{false})
query-analysis, 
yields a dilemma. For, if the interacting expert is not query-based, this binary analysis is only an approximation as there are exponentially many possible axiom-based expert answers (each query-axiom can be \emph{true}/\emph{false}/unanswered), and an exact analysis is 
impractical since exponential. 
%
(F4)~State-of-the-art methods \cite{rodler_jair-2017} can efficiently compute queries that are informative wrt.\ the minimization of \#Q under the assumption of a query-based expert, but they do neither (primarily) consider the expert's effort for query answering nor the contingency of an axiom-based expert interacting with the system.

To sum up, existing works tackle (F1)--(F4) in a way that \emph{(i)}~fault localization efficiency depends on (the behavior of) the interacting expert, \emph{(ii)}~finding of best queries might be inefficient or only approximate,
and \emph{(iii)}~optimized criteria appear to be not fully realistic.

\noindent\textbf{New Approach.}
To remedy these issues, we suggest to use so-called singleton queries instead of normal ones. A \emph{singleton query (SQ)} is a query which includes exactly one axiom (cf.\ the example query above). Albeit pretty simple, the SQ-approach 
solves all problems we discussed. 
In particular, SQs have \emph{exactly two outcomes}, entail a (necessarily) \emph{unique expert behavior} (all expert types coincide), and imply \#Ax$\,=\,$\#Q (\emph{unequivocal optimization criterion}).
Further immediate advantages are: Each computed SQ-query-axiom depends on \emph{all} so-far acquired expert inputs (\emph{better informed computations}), worst-case search costs for best SQ are less than for best normal query (\emph{smaller search space}), 
heuristic \emph{query evaluation} 
\emph{is always exact and plausible} for SQs, concepts (e.g., heuristics \cite{rodler17dx_activelearning,rodler2018ruleML,Shchekotykhin2012}, UIs \cite{schekotihin2018ontodebug}) for normal queries are directly reusable for SQs (\emph{compatibility}), with SQs there is no need to instruct experts (on how to operate for best results), to ascertain the expert's type a-priori, or to adapt algorithms to different experts (\emph{simpler computation and optimization process}), and SQs mean an \emph{equally or more informative feedback per asked axiom} (all queried axioms are indeed answered).

Hence, the SQ-technique addresses (F1)--(F3) in an elegant way, in that queries are defined as SQs (F1), which directly answers, and thus obviates the need to care about, (F2) and (F3). 
%
%
Solely, regarding (F4), there is a hitch
related to the (per-se favorable) smaller search space for SQs, in that a more sophisticated query search is required to ensure that the output is indeed an SQ.
Whereas the conception of an efficient general algorithm for SQs is an open research question, we were able to develop a polynomial time and space algorithm to find, wrt.\ a given set of diagnoses, the best SQ among those of the form $Q=\{\tax\}$ where $\tax$ is an element of the ontology.

\section{Evaluation Results and Concluding Remarks}
\label{sec:eval} 
\noindent\textbf{Evaluation.} We conducted extensive experiments on faulty real-world ontologies to study normal queries in combination with the discussed expert types and to compare 
them
with SQs. In the tests, we used only queries $Q \subseteq \mathcal{O}$ for each ontology $\mathcal{O}$.
Focus of the investigations were the discussed goals (G1)--(G3). Specifically, we examined the following questions: \emph{(Q1)}~Does the expert answering behavior make a difference (wrt.\ fault localization cost)? \emph{Answer:} For normal queries, yes. We observed significant differences (overheads of up to 140\,\%) between the distinct expert types (in terms of both \#Q and \#Ax). For SQs, trivially no, as all expert behaviors coincide. \emph{(Q2)}~Which strategy is the best to answer normal queries?
\emph{Answer:} Wrt.\ \#Ax: The pragmatist always performed superior (on avg.) to all others. Wrt.\ \#Q: It was not clear-cut, but overall the pragmatist tended to be the best as well. Hence, when relying on normal queries, we should advise experts to pursue the pragmatist answering approach.
\emph{(Q3)}~What is better, SQs or normal queries? 
\emph{Answer:} Wrt.\ \#Ax: SQs led to less expert effort in two thirds of the tested cases. Wrt.\ \#Q: Interestingly, SQs were mostly the better choice as well.
\emph{(Q4)}~Which approach (query type and answering strategy) is computationally most efficient (wrt.\ the expert's waiting time)?
\emph{Answer:} SQs. Time savings against normal queries were substantial (always larger than 80\,\%),
which can be attributed to the smaller search space.

\noindent\textbf{Conclusion.} 
Singleton queries represent an elegant solution to all discussed problems of existing query-based fault localization methods, and moreover enable a successful determination of the faulty axioms while proving more efficient on avg.\ than existing techniques in terms of both expert interaction effort and expert waiting time.

\fontsize { 9pt }{ 9pt } 
\selectfont

\end{document}